%% file: main_KK.tex
\documentclass[11pt]{article}

\usepackage[final]{acl}

\usepackage{times}
\usepackage{latexsym}
\usepackage[T1]{fontenc}
\usepackage[utf8]{inputenc}
\usepackage{microtype}
\usepackage{inconsolata}

\usepackage{float}
\usepackage{placeins}

\usepackage{graphicx}
\usepackage{caption}
\usepackage{booktabs}
\usepackage{multirow}
\usepackage{array}
\usepackage{amsmath}
\usepackage{amssymb}
\usepackage{pifont}
\usepackage{tikz}
\usetikzlibrary{arrows.meta,positioning,fit,calc}

\usepackage{xurl}   
\hypersetup{
  pdftitle={EviStreams: Human-in-the-Loop AI Data Extraction for Systematic Reviews in Medicine},
  pdfauthor={Sai Karthik Kosuri and Ankita Shashikant Bhosale and Michael Glick and Alonso Carrasco-Labra and Chris Callison-Burch},
  pdfkeywords={systematic reviews; evidence synthesis; human-in-the-loop; information extraction}
}

\usepackage{xspace}
\newcommand{\sys}{\textsc{evistreams}\xspace}

\newcommand{\fullc}{\ding{108}}
\newcommand{\halfc}{\ding{119}}

\title{\texorpdfstring{\sys}{evistreams}: Human-in-the-Loop AI Data
Extraction for Systematic Reviews in Medicine\thanks{Live demo:
\url{https://evistreams.com/demo}}}

\author{
  \textbf{Sai Karthik Kosuri}\textsuperscript{1} \quad
  \textbf{Ankita Shashikant Bhosale}\textsuperscript{1} \quad
  \textbf{Michael Glick}\textsuperscript{1} \\
  \textbf{Alonso Carrasco-Labra}\textsuperscript{1} \quad
  \textbf{Chris Callison-Burch}\textsuperscript{2} \\[4pt]
  \textsuperscript{1}Center for Integrative Global Oral Health (CIGOH), \\
  Penn Dental Medicine, University of Pennsylvania \\
  \textsuperscript{2}Department of Computer and Information Science,
  University of Pennsylvania \\[4pt]
  {\small Corresponding author: \texttt{karthik9@upenn.edu}}
}

\begin{document}
\raggedbottom
\maketitle

\begin{abstract}
Systematic reviews underpin clinical guidelines, yet their data-extraction
step is a major expert-labor bottleneck bound by a protocolized workflow: two
reviewers extract each study independently, an adjudicator resolves
disagreements, and the team keeps an auditable record of how every value was
produced. Large language models can assist with extraction, but that
assistance must fit established review protocols and preserve
reproducibility. We
present \sys, a live, open-source, no-code web platform that puts review teams
in control of AI-assisted extraction at three key stages: \emph{program
design} (a structured decomposition approved before any code
runs), \emph{field specification} (typed field definitions calibrated from a
pilot), and \emph{extracted predictions} (reviewer-blinded dual review with
adjudication). Working through a form builder, a domain expert defines typed
fields rather than prompts, runs extraction over uploaded PDFs, inspects every
value alongside the supporting passage it came from, and resolves a reviewer-blinded dual
review into an auditable consensus export. An evaluation across four clinical
corpora and three frontier model families, released with the system, shows
that extraction quality is shaped far more by the field specification than by
the choice of model. \sys is live at \url{https://evistreams.com/demo} and released
under Apache-2.0.
\end{abstract}

\input{sections/01_intro.tex}

\input{sections/02_system.tex}

\input{sections/04_setup.tex}

\input{sections/demo.tex}

\input{sections/03_related.tex}

\input{sections/06_conclusion.tex}

\input{sections/07_limitations.tex}

\clearpage
\input{sections/08_ethics.tex}

\input{sections/08b_acknowledgments.tex}

\bibliography{references}

\clearpage
\appendix
\raggedbottom
\input{sections/09_appendix.tex}

\end{document}

%% file: sections/01_intro.tex
\section{Introduction}
\label{sec:intro}

Systematic reviews inform clinical and public health practice guidelines, health-technology assessments,
and payer and regulatory decisions. Their authority rests on a demanding
protocol: two reviewers extract each study independently, an adjudicator
resolves disagreements, blinding guards against anchoring, and the team keeps
an auditable record of how every value was produced. Data extraction (the
transfer of numerical and categorical findings into structured tables) is
among the most labor-intensive steps of a systematic review
\citep{higgins2023cochrane}. Large language models can fill in much of an
extraction form, though how well varies sharply from field to field
\citep{simmons2025bespoke}, and an extraction error can be
catastrophic: a single mis-extracted sample size or effect estimate can
propagate into a pooled estimate and a clinical recommendation. Therefore we
are motivated to design a human-in-the-loop system for systematic reviews
that lets humans oversee the review, trace each value, and verify the
outputs. We investigate three key stages that put humans in control of
AI-generated extraction while meeting the evidentiary standards of the
discipline.

Our main finding is that extraction
quality is governed far less by \emph{which} model reads a paper than by
\emph{how} the task is specified. This echoes the growing observation in
clinical information extraction that ``success may increasingly hinge on the
clear articulation of objectives, rather than on singular workflow methodologies''
\citep{hein2025iterative}, and that customized, domain-tuned prompts dominate
generic ones \citep{li2025automation}, which we elevate into an architectural
commitment. We study extraction across four clinical corpora. Choosing the
model or changing the extraction pipeline changes performance by only a few
F1 points, with mixed direction across corpora. The field specification has
the largest effect: a richer, more detailed specification produces better
results.

\begin{figure}[t]
\centering
\includegraphics[width=0.9\linewidth]{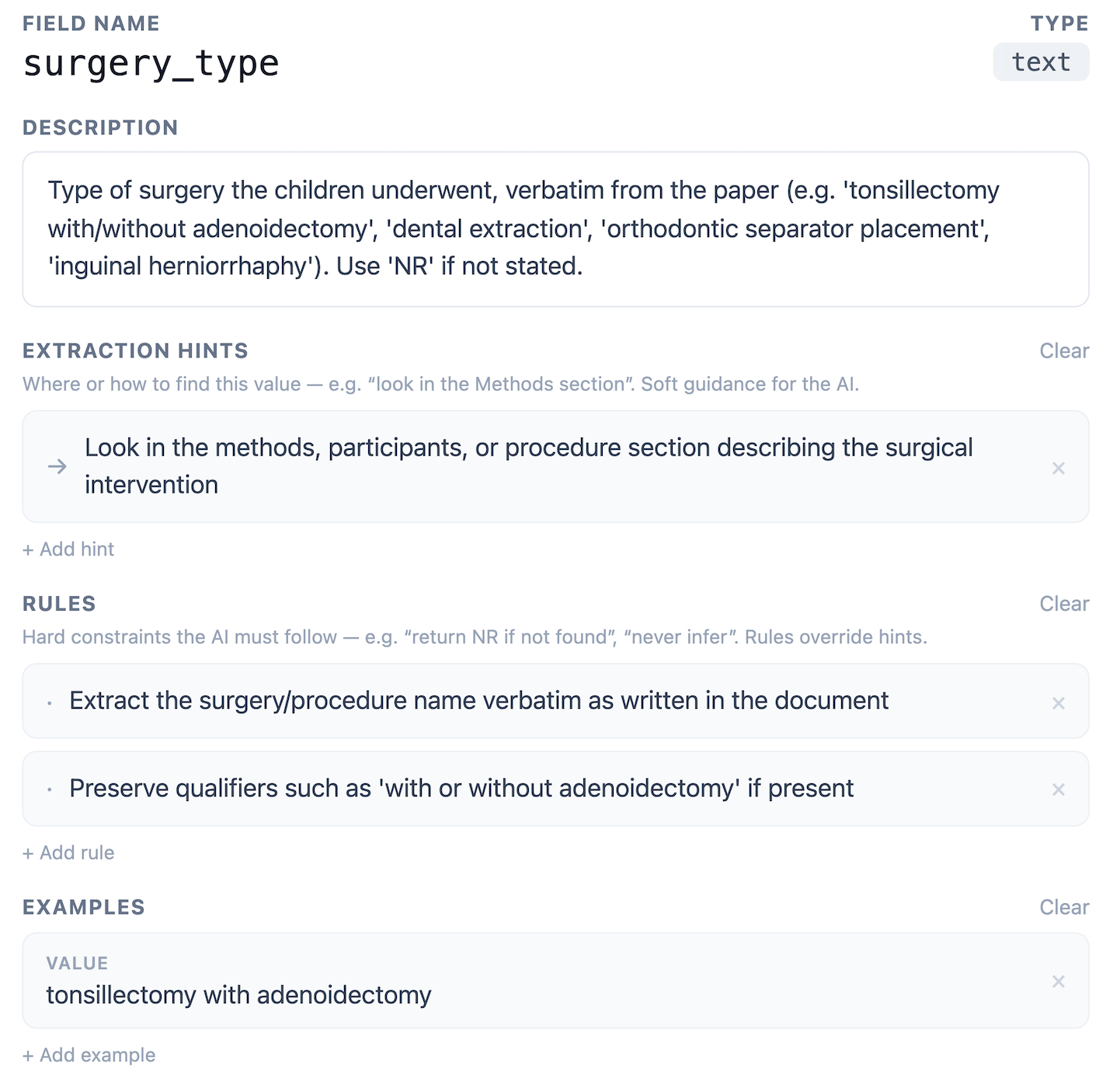}
\caption{A structured field in the form builder (\texttt{surgery\_type}):
description, extraction hints, rules, and examples, edited directly by
the reviewer.}
\label{fig:field-example}
\end{figure}

We present \sys (short for evidence streams), an open-source, live platform running at
\url{https://evistreams.com/demo} that
puts domain experts in control of AI-assisted extraction
through a no-code interface. Human review enters at three key
stages: (1)~\emph{program}, where the system proposes a plan for extracting
each field from the PDF and a reviewer approves, edits, or rejects it before
any code runs; (2)~\emph{specification}, where each field's definition is
tested on a handful of papers and the reviewer refines it based on what
comes back; and (3)~\emph{predictions}, where every extracted value is
checked by two independent reviewers, and a third resolves any
disagreement. Figure~\ref{fig:field-example} shows how the domain expert controls
extraction by adding hints, rules, and examples.

Every AI-extracted value is paired with the
verbatim quote it was extracted from, which reviewers can inspect. Unlike
prior work, which lets users correct only the final output, \sys gives many
ways to control how the system behaves: from the extraction plan to the
field specification to the final adjudication step.

%% file: sections/02_system.tex
\section{The \sys System}
\label{sec:system}

We designed \sys for the ADA Living Guidelines Program, a collaboration
between the American Dental Association and our lab, the Center for
Integrative Global Oral Health (CIGOH), that keeps oral health guidelines
updated as new evidence comes in
\citep{carrascolabra2026living}. Our team uses \sys end to end: a
methodologist designs a structured extraction form and approves the
pipeline the system proposes for it, then reviewers run extraction over
the uploaded PDFs and adjudicate the results, with human review entering
at three key stages (Figure~\ref{fig:architecture}).

\subsection{Extraction pipeline}
\label{sec:methods-pipeline}
\begin{figure}[H]
\centering
\includegraphics[width=\linewidth]{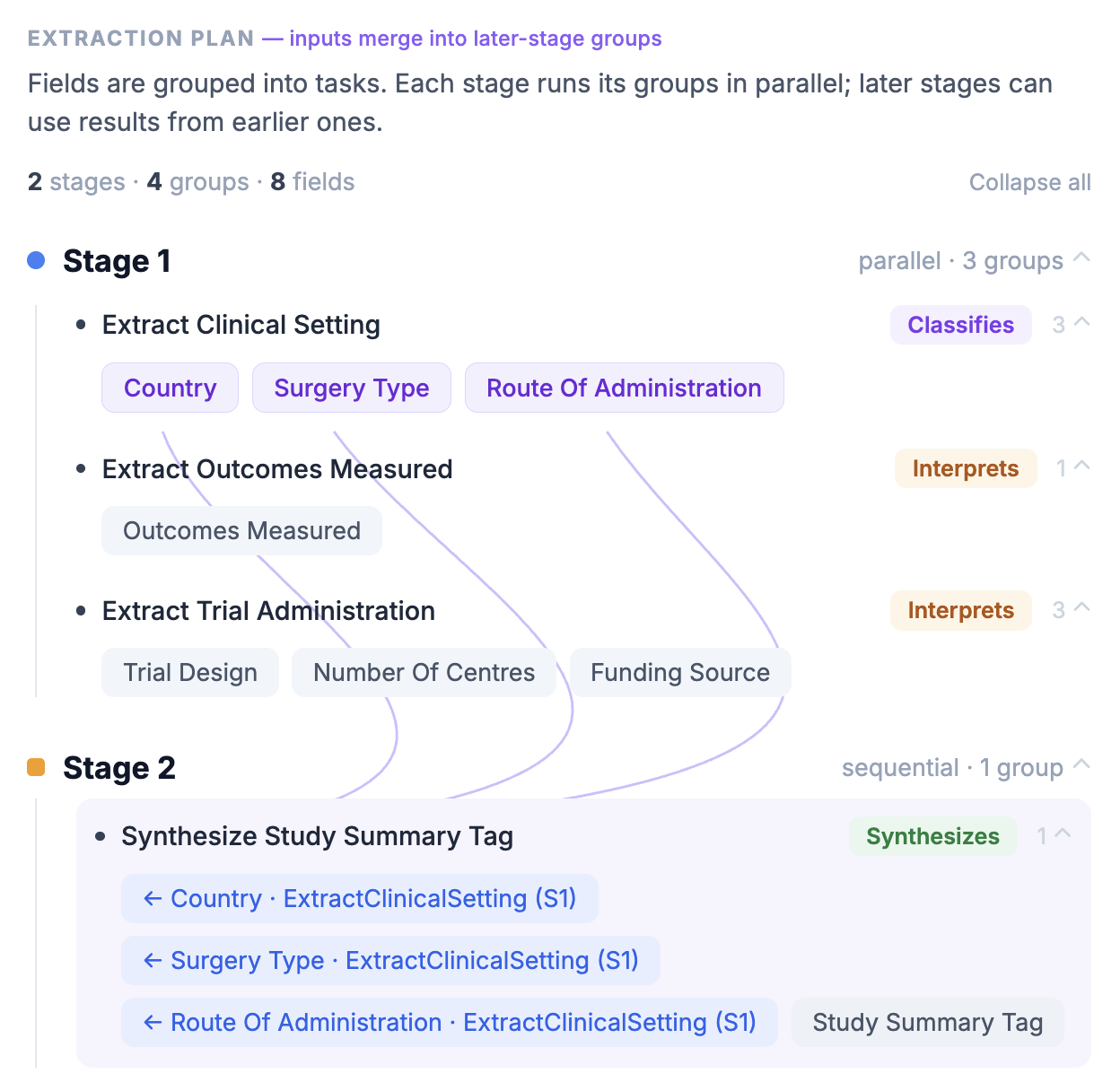}
\caption{An extraction plan: fields are grouped into tasks, and each stage
runs its groups in parallel; later stages can use results from earlier
ones. Here Stage~1 extracts clinical setting, outcomes, and trial
administration fields in parallel; Stage~2 synthesizes a study summary tag
from three of Stage~1's outputs.}
\label{fig:extraction-plan}
\end{figure}

A methodologist designs a \emph{form}, giving only a name, a description,
and optionally an example for each field. The system groups related fields
into stages and proposes this grouping as a plan (Figure~\ref{fig:extraction-plan})
for the reviewer to
approve: never executable code. Once approved, \sys\footnote{Backend:
\url{https://github.com/karthik-strikes/evistream_backend}. Frontend:
\url{https://github.com/karthik-strikes/evistream_frontend}. Released
under Apache-2.0.} adds hints and rules
to each field and turns the plan into extraction code automatically: a
fixed compiler maps the approved plan to predefined DSPy components; no
LLM writes this code. Separately, uploaded PDFs are stored in S3 and
converted to Markdown using Datalab~\citep{datalab2024}.

\begin{figure}[t]
\centering
\includegraphics[width=\linewidth]{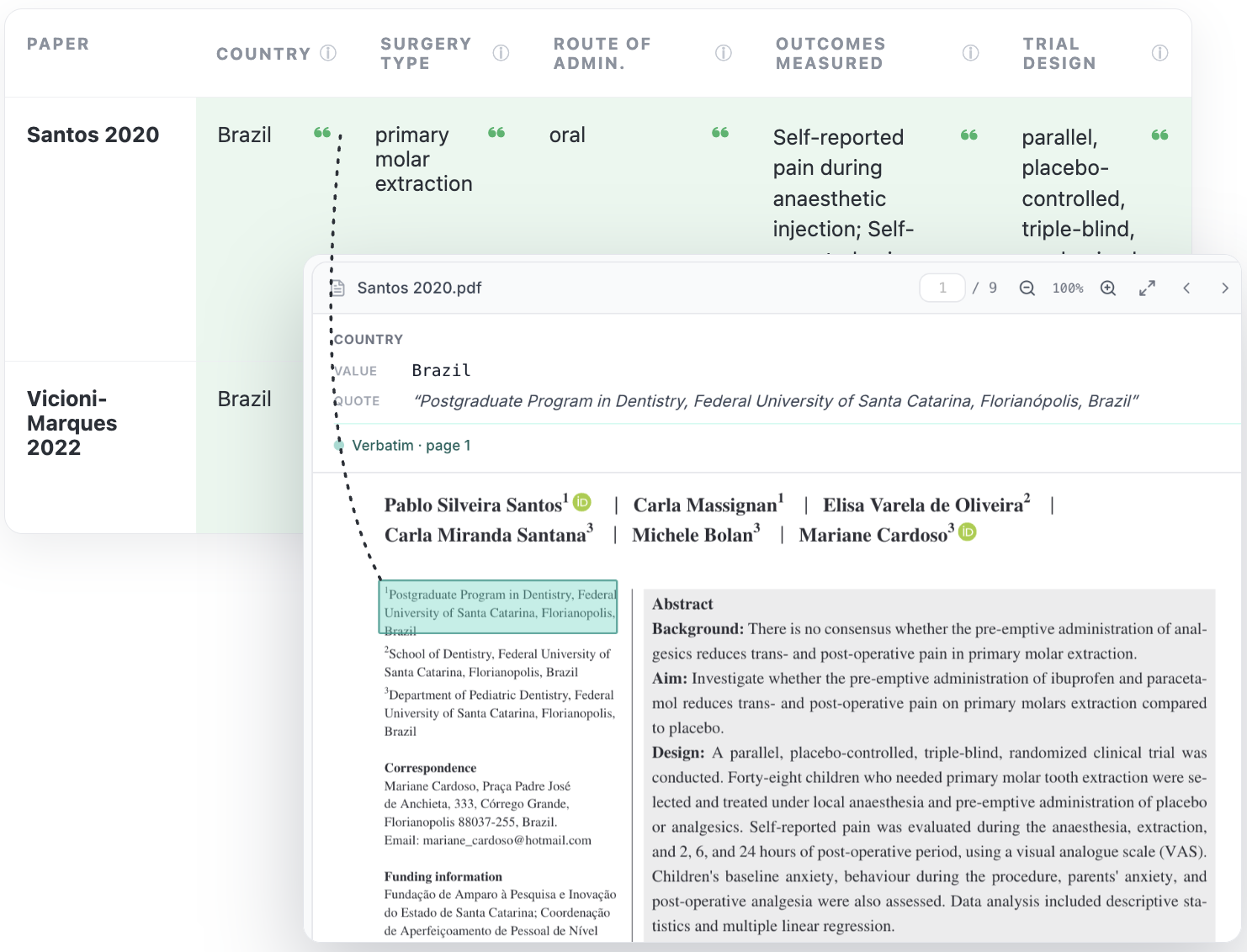}
\caption{Run: every extracted value can be expanded to its verbatim source
quote and page in the original PDF, so a reviewer can verify a value (here
\texttt{country}: ``Brazil'') against exactly the sentence it came from.}
\label{fig:run}
\end{figure}

Once fields are enriched, each group from the plan becomes a DSPy
signature \citep{khattab2023dspy}: fields in the same group share one
Chain-of-Thought signature \citep{wei2022cot}, and independent groups run
at the same time. A DSPy module then runs each signature to extract the
data. We skip DSPy's automatic prompt optimizers
\citep{opsahlong2024miprov2,agrawal2025gepa}: at our scale (5--20 example
papers), we prioritize direct expert editing over automatic prompt
optimization \citep{li2025automation}. Every value comes back with its source text, in
the same format across every model (Figure~\ref{fig:run}).

\subsection{Structured field specification}
\label{sec:methods-field}
Each field's specification has four elements (Figure~\ref{fig:field-example},
from the introduction): a \textbf{description} of the concept (always
present, user-owned and never rewritten); \textbf{hints} locating the
evidence within a paper; \textbf{rules} imposing hard output constraints
(e.g.\ ``return the analyzed $N$, not the enrolled $N$''); and
\textbf{examples} anchoring the expected value shape. Hints and rules are
filled in automatically: a LangGraph~\citep{langgraph2024} chain runs a
prompt over the field's description and any user-given examples to
generate them, consistent with what the user wrote.

Splitting \emph{what} a field means from \emph{how} to find it
lets a methodologist localize an error to just one element. These elements
are also what pilot calibration edits later (\S\ref{sec:methods-hitl}); the
reviewer refines them after seeing pilot results.

\subsection{Table extraction}
\label{sec:methods-table}
Medical-study tables record one row per \textbf{arm} (the different
drugs or interventions a study tests for a condition, e.g., ibuprofen
versus an alternative drug), or one row per (arm, follow-up, outcome)
triplet. Naive LLM extraction on these tables often truncates the table,
drifts in meaning, or hallucinates columns that were never asked for.

We avoid this by decomposing the extraction \citep{khot2023decomposed}: instead
of reading a table positionally, we first extract only a few \emph{anchor
columns} that name each row (e.g., drug name and dose), then fan out one
call per row to fill in the rest. For example, a 10-column table repeated
across 400 papers easily loses columns under naive extraction; anchoring
each row to its drug and dose first, then filling in the remaining cells
per row, is designed to reduce row and column drift, since each call
already knows which row it belongs to.

\subsection{Human review at three key stages}
\label{sec:methods-hitl}
\textbf{Program review} (\S\ref{sec:methods-pipeline},
Figure~\ref{fig:extraction-plan}) is implemented as a
LangGraph~\citep{langgraph2024} workflow: the proposed plan is a directed
acyclic graph (DAG) of extraction sub-tasks with field-level dependencies,
validated mechanically for cycles and for missing or duplicated field
assignments (regenerating with structured feedback on failure), and paused
at an \texttt{interrupt\_before} checkpoint until the reviewer approves,
edits, or rejects it.

\textbf{Specification calibration} occurs after a form is
active: the reviewer tests the form on a handful of papers (typically
3--5), sees the results, and re-edits the fields (Figure~\ref{fig:field-example})
based on what comes back.
These corrections update the corresponding signatures at runtime: any
change to a field's hints, rules, or description is reflected directly in
its prompt.

\begin{figure}[H]
\centering
\includegraphics[width=\linewidth]{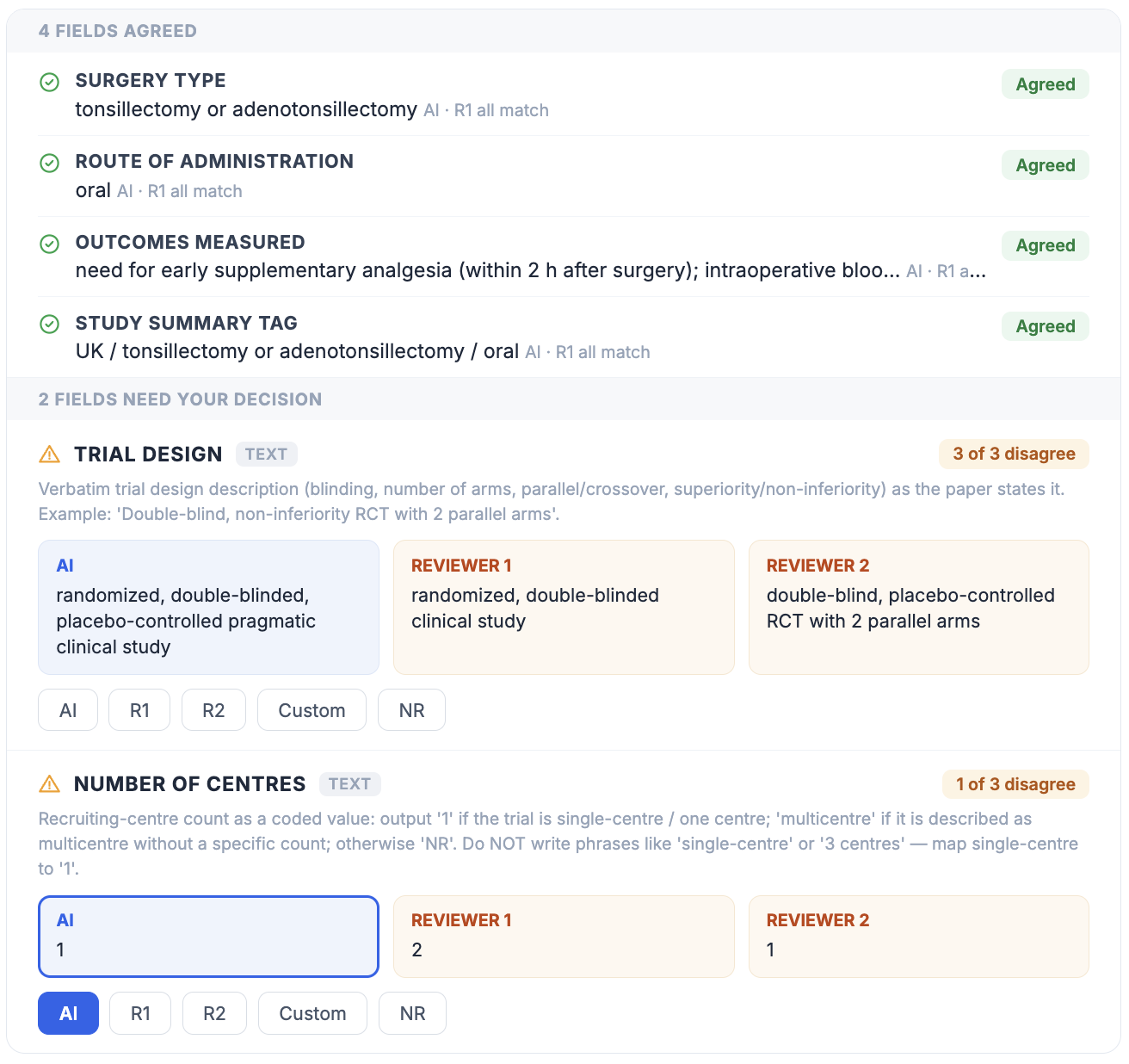}
\caption{Adjudication: fields where the AI and both reviewers already
agree are marked \emph{Agreed}; fields with a disagreement (here
\texttt{trial\_design} and \texttt{number\_of\_centres}) show the AI's and
both reviewers' entries side by side for the adjudicator to pick or
correct.}
\label{fig:adjudication}
\end{figure}

\textbf{Adjudication}\footnote{Screencast ($\le$2.5~min):
\url{https://www.youtube.com/watch?v=j1Gxyu0KitU}.} occurs after extraction: two
reviewers independently complete the same form on the same paper under
reviewer blinding (each sees the AI pre-fill but not the other's entries).
Field-level disagreements are computed on demand, and an adjudicator
resolves each conflict (Figure~\ref{fig:adjudication}). For every field,
the system links the AI's
pre-fill, both reviewers' entries, and the adjudicator's decision
together, so the full history of how a value moved from AI guess to final
answer can be looked up later. When a reviewer changes an AI value they
attach their own supporting quote, so every value still shows where in the
paper it came from: the system keeps the AI's value, the reviewer's
replacement, and a pointer into the PDF for each. Reviewer blinding keeps one
reviewer's entry from anchoring the other, preserving the independent review
that systematic-review protocols require; it does not blind either reviewer
to the AI pre-fill itself.

\paragraph{Implementation.} \sys is a Next.js/FastAPI stack with Celery
workers, Redis-backed live WebSocket progress, Supabase (Postgres) storage,
Markdown ingestion (Datalab), and a multi-model fallback chain over the three
frontier families.

%% file: sections/04_setup.tex
\section{Evaluation}
\label{sec:setup}\label{sec:results}
We evaluate \sys on four clinical datasets, measure the differences among
three frontier models, and analyze the errors that occur. The choice of
frontier model does not dramatically affect quality: all three perform
similarly. The field specification drives the largest variance in
quality.

\paragraph{Setup.}
\label{sec:methods-corpora}\label{sec:methods-scoring}\label{sec:methods-ablations}
We score results with a field-typed harness, released publicly alongside
\sys and implemented independently from the extraction pipeline to reduce
evaluation coupling. We use four clinical corpora whose ground truth was drawn, as reported,
from completed, peer-reviewed systematic reviews in which two review authors
extracted every study independently and resolved disagreements by consensus,
the same protocol \sys supports: \textbf{oral cancer}
(three reviews of oral cancer diagnostic adjuncts~\citep{verdugopaiva2026cytology,bhosale2026vitalstaining,%
urquhart2026lightbased}; 43 diagnostic-accuracy studies, four forms,
47 fields), \textbf{antibiotic prophylaxis}
(CD010266~\citep{brignardellopetersen2015antibiotic}; 10
trials), \textbf{periodontitis} (CD004714~\citep{simpson2022periodontitis}; 29 trials, outcomes per
arm\,$\times$\,time point), and \textbf{ibuprofen} (CD015432~\citep{pessano2024ibuprofen}; 43 trials); all three model families were run on every corpus
(Appendix~\ref{app:coverage})\footnote{Extraction models: Claude
Sonnet~4.6~\citep{anthropic2026sonnet46}, GPT-5.5~\citep{openai2026gpt55}, and
Gemini~3.1~Pro~\citep{googledeepmind2026gemini31pro} (preview); the free-text judge is
Claude Sonnet~4.6. We score every study whose full-text PDF was
available: 10 of 11 antibiotic-prophylaxis trials and 29 of 35
periodontitis studies (the rest were excluded for PDF unavailability);
ibuprofen is complete at 43 of 43.}.

A field's initial definition comes from
the review protocol, which sets out the PICO definitions, the eligibility
criteria and the outcome definitions before anyone extracts anything, and which
the reviewers who produced our reference data worked from as well. Its hints,
rules and examples were then refined on pilot papers drawn from the same corpus
we later score against.

Each field is scored by a type-matched strategy,
and record-level forms are aligned by a globally optimal Hungarian assignment
over row-identity keys before scoring, so a missed arm counts against recall
and a spurious one is an over-extraction. We report micro- and macro-F1,
which agree within $\sim$0.02 on most forms (up to $\sim$0.06 on the hardest
oral-cancer forms).

We test the model, the pipeline decomposition, and the field specification
one at a time, holding the other two fixed. We score against the
reference data, though this ground truth can itself
contain reviewer conventions or occasional errors. The
free-text judge
(Claude Sonnet~4.6) is itself one of the three extraction models, a
potential circularity readers should weigh when interpreting free-text
scores. That judge decides 34\% of scored comparisons; the rest are settled
by deterministic numeric or normalised-string matching.\label{sec:results-judge}

\subsection{Production quality and robustness}
\label{sec:results-scorecard}\label{sec:results-model}\label{sec:results-decomp}
Table~\ref{tab:corpora} reports overall micro-F1 per corpus for all three
frontier families through the identical pipeline and specifications (per-form
detail in Appendix~\ref{app:percorpus}). Quality varies little across the
families tested (Claude and Gemini within $0.03$ micro-F1 on every corpus, GPT trailing
by up to $\sim$0.05 on antibiotic and periodontitis), and is bounded far more by
the form and its source text than by the model (the between-corpus range,
$0.741$--$0.898$, dwarfs the between-model one). Oral cancer, our most complex
corpus (four forms, 47 fields), is consistently hardest across all three
models; antibiotic prophylaxis is consistently easiest, tracking form
complexity rather than any one model's weakness. With specification and pipeline
fixed, the provider can thus be chosen on cost, latency, or governance rather
than accuracy. Collapsing the multi-signature pipeline into a single hand-tuned
prompt changes overall F1 by at most $0.036$ with mixed direction (the one
sizable swing, antibiotic tables, is GPT-specific: $\Delta=-0.096$ vs.\ $-0.005$
for Claude and Gemini), so decomposition provides governable structure, not
consistent accuracy gains (Appendix~\ref{app:decompfull}).

\begin{table}[t]
\centering
\small
\setlength{\tabcolsep}{4pt}
\begin{tabular}{lrccc}
\toprule
Corpus & Studies & Gemini & GPT & Claude \\
\midrule
Ibuprofen     & 43 & 0.842 & 0.845 & \textbf{0.851} \\
Oral cancer   & 43 & 0.741 & 0.758 & \textbf{0.763} \\
Antibiotic    & 10 & 0.896 & 0.851 & \textbf{0.898} \\
Periodontitis & 29 & 0.827 & 0.799 & \textbf{0.846} \\
\bottomrule
\end{tabular}
\caption{Overall micro-F1 per corpus, three frontier model families through
the same pipeline with identical form specifications (best per row in bold).
Macro- and micro-F1 differ by at most $0.03$ at the corpus level; per-form
macro-F1 detail in Appendix~\ref{app:percorpus}.}
\label{tab:corpora}
\end{table}

\subsection{Field specification is the largest observed lever}
\label{sec:results-spec}
Across the four calibrated corpora and three tested model families, richer
field specifications improved macro-F1 in 40 of 51 model--form comparisons
($78\%$), with a mean $\Delta F_1$ of $0.049$ and a median of $0.014$ (full
per-form ablation in Appendix~\ref{app:specfull}). The mean is strongly
influenced by the oral-cancer \texttt{index\_test} form: removing hints,
rules, and examples reduced its macro-F1 from $0.724$ to
$0.247$ for Claude, with similarly large declines for Gemini and GPT.
Excluding this form, the mean improvement was $0.023$, and the full
specification remained better in 37 of 48 comparisons ($77\%$). Thus,
richer specifications usually provide modest gains but can become decisive
for fields requiring disambiguation, source localization, or non-obvious
extraction conventions. Model and decomposition effects, reported above,
were comparatively small and mixed in direction.

\subsection{Error profile}
\label{sec:results-errors}\label{sec:results-synth}

\input{sections/error_table.tex}

We reviewed and classified every scored discrepancy in 52 studies (all 10
antibiotic trials; 15/43 oral-cancer, 15/43 ibuprofen, 12/29 periodontitis,
each of the latter three by set-cover over the lowest-scoring fields): an
LLM-assisted first-pass classification, followed by manual verification of
every label by an expert systematic-review methodologist.
Table~\ref{tab:errorprofile} summarizes the discrepancy taxonomy and
prevalence: of $1{,}003$ discrepancies, two-thirds are \textbf{methodological}
(ground-truth, convention, ambiguity, or scoring-side), not model failures;
under a third are \textbf{genuine model errors}, concentrated in dense
multi-arm, multi-table reporting; the rest are unadjudicable. Among genuine
model errors, omissions and context-assignment mistakes were both common.
Missing rows and false-NR omissions accounted for 138 discrepancies, while
160 involved incorrect arms, time points, cohorts, denominators, or
regimens. Pilot calibration is intended to reduce specification-driven
misses, while dual review and adjudication provide a final check on
residual errors.
Detailed definitions and causes for each
subtype are provided in Appendix~\ref{app:errorfull}.

%% file: sections/error_table.tex
\begin{table}[t]
\centering
\small
\setlength{\tabcolsep}{4pt}
\begin{tabular}{@{}p{5.4cm} r@{}}
\toprule
Discrepancy type & Share \\
\midrule
\textbf{Methodological} (ground-truth/protocol side) & 68\% \\
\midrule
\multicolumn{2}{@{}l}{\textbf{Genuine model errors} (30\%)}\\
\quad Missing record or outcome row & 7\% \\
\quad Stated-value omission (\texttt{false NR}) & 6\% \\
\quad Wrong arm, table, or time point & 6\% \\
\quad Cohort or adjacent-column confusion & 5\% \\
\quad Denominator or document-context error & 4\% \\
\quad Arm or regimen conflation & 1\% \\
\midrule
\textbf{Uncertain/unadjudicable} & 2\% \\
\bottomrule
\end{tabular}
\caption{Taxonomy of $1{,}003$ reviewed discrepancies from an error-enriched
sample of 52 studies. Percentages are shares of reviewed discrepancies
(rounded; subtypes need not sum to the group total), not extraction error
rates. Detailed definitions and causes appear in
Appendix~\ref{app:errorfull} (Table~\ref{tab:errorfull}).}
\label{tab:errorprofile}
\end{table}

%% file: sections/demo.tex
\section{Demonstration}
\label{sec:demo}

\paragraph{Demo walkthrough.} In the reviewer demo, a methodologist first
uploads a batch of PDFs, which \sys converts to Markdown via
Datalab~\citep{datalab2024} in the background while the methodologist keeps
working. The methodologist then creates a form, naming each field with a
description and an optional example; \sys proposes a decomposition that
groups related fields into parallel extraction stages
(Figure~\ref{fig:extraction-plan}), and the methodologist reviews and
approves the plan before any code runs.

\sys then runs a pilot study, extracting the form over 3--5 papers so the
methodologist can check each AI value against the expected value and refine
hints, rules, or examples on any field that came back wrong; edits save
live and are reflected in the next pilot run.

Once the pilot output matches expectations, the methodologist runs the
approved form over the remaining documents, and every extracted value is
shown alongside the supporting passage it was drawn from
(Figure~\ref{fig:run}). Two reviewers then carry out manual extraction:
each independently completes the same form on the same paper under
reviewer blinding, accepting or correcting the AI pre-fill. \sys surfaces their
field-level disagreements, and an adjudicator resolves each conflict into
the consensus export (Figure~\ref{fig:adjudication}), which records the
AI suggestion, both reviewer entries, the adjudicated value, and
inter-rater agreement.

\paragraph{Target audience.} Systematic-review teams: clinical experts who
extract data, and methodologists who need dual review, adjudication,
and inter-rater statistics for Cochrane-style protocols
\citep{higgins2023cochrane}. A secondary audience
is NLP researchers, for whom the per-field decision chain (AI pre-fill, R1, R2,
adjudicator) is a reusable data asset.

%% file: sections/03_related.tex
\section{Related Work}
\label{sec:related}

\paragraph{Systems for evidence synthesis.}
\textsc{AutoForest} \citep{autoforest2026} targets a different
endpoint (Cochrane forest-plot generation from PDFs) with interactive
correction of ICO elements, extracted values, and synthesis outputs.
\textsc{ScheMatiQ} \citep{levy2026schematiq} discovers a schema from a research
question plus a document collection and supports expert revision of the schema
and the extracted table; it is our closest general-purpose sibling.

Established review
tools \citep{marshall2017robotreviewer,iyer2025docspiral}
and annotation platforms \citep{pei2022potato,klie2018inception} place the
human loop at the prediction level only, correcting outputs rather than the
program that produces them.

Table~\ref{tab:features} compares the extraction systems directly. Several
let a user define the extraction schema, and most let a reviewer correct
extracted values, and the commercial platforms carry blinded dual review with
adjudication. \sys differs in one place outright, review at the program level
before any extraction runs, and in combining that with specification
calibration and blinded dual review in a single workflow.

\begin{table}[t]
\centering
\small
\setlength{\tabcolsep}{2pt}
\begin{tabular}{l ccccccc}
\toprule
System & UDS & AIX & SRC & OUT & PRG & CAL & DRA \\
\midrule
\textsc{ScheMatiQ}     & \fullc & \fullc & \fullc & \fullc & ---    & \fullc & ---    \\
\textsc{AutoForest}    & \halfc & \fullc & \halfc & \fullc & ---    & ---    & ---    \\
\textsc{ROBoto2}       & ---    & \fullc & \fullc & \fullc & ---    & ---    & ---    \\
Elicit                 & \fullc & \fullc & \fullc & \fullc & ---    & \halfc & \halfc \\
Covidence              & \fullc & \halfc & \halfc & \fullc & ---    & ---    & \fullc \\
DistillerSR            & \fullc & \fullc & \fullc & \fullc & ---    & ---    & \fullc \\
\midrule
\sys                   & \fullc & \fullc & \fullc & \fullc & \fullc & \fullc & \fullc \\
\bottomrule
\end{tabular}
\caption{Extraction-assistance capabilities across systems:
\textsc{ScheMatiQ} \citep{levy2026schematiq}, \textsc{AutoForest}
\citep{autoforest2026}, \textsc{ROBoto2} \citep{hevia2025roboto2}, and the
commercial platforms Elicit, Covidence and DistillerSR, whose capabilities
were verified against their official documentation
\citep{elicit2026,covidence2026ai,covidence2026consensus,distillersr2026genai,%
distillersr2016extraction}; \textsc{Trialstreamer}
\citep{nye2020trialstreamer} and \textsc{ClinicalTrialsHub}
\citep{park2026clinicaltrialshub} extract fixed elements rather than a
review-defined form and are not compared here.
\textbf{UDS}: user-defined schema;
\textbf{AIX}: AI-generated pre-fill;
\textbf{SRC}: value-level source grounding (each value linked to its verbatim
passage);
\textbf{OUT}: reviewer correction of extracted outputs;
\textbf{PRG}: program-level review (extraction plan approved before it runs);
\textbf{CAL}: specification calibration loop (pilot results $\rightarrow$ spec
edits with immediate effect);
\textbf{DRA}: reviewer-blinded dual review with adjudication carried to the
export.
\fullc\ supported; \halfc\ partial (\textsc{AutoForest}: control over ICO
elements only, and free-text extraction rationales without links to source
locations; Covidence: field suggestions rather than form pre-fill, and a
supporting quote per suggestion without a link to its location; Elicit:
iteration over free-text column prompts rather than structured field elements,
and dual review described for screening rather than extraction);
--- not described in the cited work or documentation.
Systems target different endpoints and corpora, and no shared extraction
benchmark exists; we therefore compare capabilities, and release our evaluation
harness and four corpora to enable future head-to-head comparison.}
\label{tab:features}
\end{table}

\paragraph{Specification, decomposition, and data extraction.}
Our thesis turns a growing observation (that objective articulation
can matter more than model choice
\citep{hein2025iterative,li2025automation}) into a structured, editable control
surface tested across models. Concurrent clinical information-extraction systems have also explored modular
or agent-based decomposition
\citep{jeeballah2026multiagent,hart2026dualvalidation}.

On the data side, \textsc{VaxScope}
\citep{ilgen2026vaxscope} benchmarks review-level extraction with a fixed
taxonomy, whereas \sys exposes user-defined structured fields in a live loop. We
adopt a variant of the error taxonomy of \citet{simmons2025bespoke}; the
per-field extractor is a Chain-of-Thought module \citep{wei2022cot} compiled
through DSPy \citep{khattab2023dspy}, and our anchor-column table extraction
(\S\ref{sec:methods-table}) is a domain
instance of decomposed prompting \citep{khot2023decomposed,wu2025select}.

We
therefore do not claim novelty in pipeline decomposition or source-grounding
in themselves (both appear in the systems above), but in unifying human
review over a structured, editable specification, and
in showing empirically that, across our corpora, the field specification is
the largest and most consistent control surface for quality.

%% file: sections/06_conclusion.tex
\section{Conclusion}
\label{sec:conclusion}

\sys is a human-in-the-loop platform for conducting systematic reviews via
human--AI collaboration. It is made up of several parts: the extraction
\emph{program} (approved by users before it runs), the \emph{specification}
(iteratively refined by letting the user run test questions), and the
\emph{extractions} (validated through reviewer-blinded dual review). Our evaluation
shows, across four clinical corpora and three frontier model families, that
extraction quality is robust to model and pipeline decomposition, and that
the structured field specification drives the biggest variance in quality.
Expert effort is therefore best spent on the specification, with
adjudication the second-most-useful place for human intervention to catch
residual errors. Because a field's specification can be edited and take
effect immediately, without regenerating the pipeline, the lever our
evaluation identifies as most valuable is also the cheapest one to pull. An
extraction workflow that is expert-controlled and auditable is intended to
reduce transcription work and supports systematic-review workflows.

%% file: sections/07_limitations.tex
\section*{Limitations}
\label{sec:limitations}

\paragraph{Human agreement.} Inter-annotator agreement was not measured when
the reference data was created (\S\ref{sec:setup}).

\paragraph{Specification calibration.} We refined specifications using pilot
papers from the same reviews used for evaluation. Our results therefore measure
the benefit of review-specific, calibrated specifications. Performance when
specifications are developed for a new review without prior calibration remains
untested.

\paragraph{LLM-judge overlap.} The free-text judge shares a provider family
with one extraction model (\S\ref{sec:results-judge}), creating potential
evaluator dependence that we do not quantify.

\paragraph{Corpus scope.} We evaluated on four corpora totalling 125 studies.
All four come from completed, peer-reviewed systematic reviews, and we chose
them to differ in the kind of question they ask: diagnostic accuracy, surgical
prophylaxis, treatment of a chronic disease, and relief of post-operative pain. We are extending the evaluation to reviews from further
areas to see whether the field specification is still the biggest lever there.

%% file: sections/08_ethics.tex
\section*{Ethics Statement}
\label{sec:ethics}

\sys is a research tool for evidence synthesis, not a medical device, and
its outputs are not clinical advice; this work adheres to the ACL Code of
Ethics and the ACM Code of Ethics and Professional Conduct. Its central design property is that no AI-extracted value enters a
final dataset without human review: the dual-review and adjudication
workflow (\S\ref{sec:system}) enforces verification structurally rather
than relying on user diligence.

The evaluation corpora are published, peer-reviewed articles; no
patient-level or identifiable data is processed. Users upload PDFs they are
licensed to access; uploaded text is sent to commercial LLM APIs (Anthropic,
OpenAI, Google) subject to those providers' data-handling terms. The hosted
instance restricts project data to invited team members via role-based
access control.

Automated extraction errors could propagate into clinical guidance if left
unchecked; we report the error analysis openly
(\S\ref{sec:results-errors}), distinguishing genuine model errors from
ground-truth and annotation-convention discrepancies, and the per-field
decision chain records every step from AI suggestion to human-approved
value.

%% file: sections/08b_acknowledgments.tex
\section*{Acknowledgments}

This research was developed with funding from the Defense Advanced Research
Projects Agency's (DARPA) SciFy program (Agreement No.\ HR00112520300) and is
based upon work supported in part by the Office of the Director of National
Intelligence (ODNI), Intelligence Advanced Research Projects Activity (IARPA),
via 56000026C0019 (the BENGAL program). The views and conclusions contained
herein are those of the authors and should not be interpreted as necessarily
representing the official policies, either expressed or implied, of DARPA,
ODNI, IARPA, the Department of Defense, or the U.S.\ Government. The U.S.\
Government is authorized to reproduce and distribute reprints for governmental
purposes notwithstanding any copyright annotation therein.

%% file: sections/09_appendix.tex
\section{Ablation and Model Coverage}
\label{app:coverage}

All three model families (Claude, GPT, Gemini) were run through the
production pipeline, the single-prompt arm (decomposition ablation), and the
description-only arm (specification ablation) on all four corpora.

\section{Per-Corpus Production Results}
\label{app:percorpus}

Table~\ref{tab:main} gives the per-form macro-F1 for all four corpora behind
the main-text corpus-level micro-F1 (Table~\ref{tab:corpora}).
Interventions and outcomes are record-level (one row per arm, or per
arm\,$\times$\,time point).

\begin{table}[H]
\centering
\small
\setlength{\tabcolsep}{2.5pt}
\begin{tabular}{llccc}
\toprule
Corpus & Form & Gemini & GPT & Claude \\
\midrule
\multirow{6}{*}{Ibuprofen}
 & \texttt{study\_char.}      & 0.790 & 0.827 & \textbf{0.829} \\
 & \texttt{patient\_pop.}     & 0.920 & \textbf{0.938} & 0.933 \\
 & \texttt{interventions}     & 0.892 & 0.900 & \textbf{0.915} \\
 & \texttt{continuous\_out.}  & \textbf{0.795} & 0.772 & 0.765 \\
 & \texttt{dichotomous\_out.} & 0.774 & 0.754 & \textbf{0.792} \\
\cmidrule(l){2-5}
 & Overall & 0.834 & 0.838 & \textbf{0.847} \\
\midrule
\multirow{5}{*}{Oral cancer}
 & \texttt{study\_char.}    & 0.721 & \textbf{0.736} & 0.730 \\
 & \texttt{patient\_pop.}   & 0.788 & 0.788 & \textbf{0.819} \\
 & \texttt{reference\_std.} & 0.682 & 0.687 & \textbf{0.695} \\
 & \texttt{index\_test}     & 0.679 & 0.721 & \textbf{0.724} \\
\cmidrule(l){2-5}
 & Overall & 0.718 & 0.733 & \textbf{0.742} \\
\midrule
\multirow{5}{*}{Antibiotic}
 & \texttt{study\_char.}      & \textbf{0.921} & 0.907 & 0.919 \\
 & \texttt{patient\_pop.}     & 0.862 & \textbf{0.880} & \textbf{0.880} \\
 & \texttt{interventions}     & 0.920 & 0.853 & \textbf{0.922} \\
 & \texttt{dichotomous\_out.} & \textbf{0.874} & 0.746 & 0.862 \\
\cmidrule(l){2-5}
 & Overall & 0.894 & 0.847 & \textbf{0.896} \\
\midrule
\multirow{5}{*}{Periodontitis}
 & \texttt{study\_char.}     & \textbf{0.878} & 0.802 & 0.859 \\
 & \texttt{patient\_pop.}    & 0.777 & 0.787 & \textbf{0.815} \\
 & \texttt{interventions}    & 0.819 & 0.790 & \textbf{0.881} \\
 & \texttt{continuous\_out.} & \textbf{0.830} & 0.807 & 0.822 \\
\cmidrule(l){2-5}
 & Overall & 0.826 & 0.797 & \textbf{0.844} \\
\bottomrule
\end{tabular}
\caption{Per-form macro-F1, three model families through the same pipeline
with identical form specifications, across all four corpora (best per row
in bold; ties bolded on both). Corpus-level micro-F1 is in
Table~\ref{tab:corpora}. Periodontitis is scored on the canonical 29-field
set (OPTIONAL and structural row-matching fields excluded).}
\label{tab:main}
\end{table}

\section{Specification Ablation, Full Detail}
\label{app:specfull}

Table~\ref{tab:specfull} gives the per-form specification ablation ($\Delta
F_1$\,=\,full\,$-$\,description-only) for all three model families, the
finding summarized in \S\ref{sec:results-spec}. Full-pipeline per-form
macro-F1 is in Table~\ref{tab:main} (Appendix~\ref{app:percorpus}). The
oral-cancer \texttt{index\_test} recall collapse discussed in the main text
reproduces across all three models (Claude $+0.476$, Gemini $+0.489$, GPT
$+0.433$).

\begin{table}[H]
\centering
\footnotesize
\setlength{\tabcolsep}{1.5pt}
\begin{tabular}{llccc}
\toprule
Corpus & Form & Cld. & Gem. & GPT \\
\midrule
\multirow{4}{*}{Antibiotic}
 & \texttt{study\_char.}      & $+0.012$ & $+0.029$ & $+0.014$ \\
 & \texttt{patient\_pop.}     & $\phantom{+}0.000$ & $-0.018$ & $+0.031$ \\
 & \texttt{interventions}     & $+0.029$ & $+0.026$ & $+0.005$ \\
 & \texttt{dichotomous\_out.} & $+0.014$ & $+0.037$ & $+0.092$ \\
\midrule
\multirow{4}{*}{Periodont.}
 & \texttt{study\_char.}    & $-0.027$ & $+0.013$ & $+0.036$ \\
 & \texttt{patient\_pop.}   & $+0.014$ & $+0.002$ & $+0.008$ \\
 & \texttt{interventions}   & $+0.065$ & $+0.116$ & $+0.307$ \\
 & \texttt{continuous\_out.}& $+0.004$ & $+0.022$ & $+0.058$ \\
\midrule
\multirow{5}{*}{Ibuprofen}
 & \texttt{study\_char.}      & $-0.011$ & $-0.030$ & $+0.010$ \\
 & \texttt{patient\_pop.}     & $-0.001$ & $+0.002$ & $+0.010$ \\
 & \texttt{interventions}     & $-0.014$ & $+0.041$ & $+0.018$ \\
 & \texttt{continuous\_out.}  & $-0.033$ & $+0.001$ & $+0.006$ \\
 & \texttt{dichotomous\_out.} & $+0.044$ & $\phantom{+}0.000$ & $-0.050$ \\
\midrule
\multirow{4}{*}{Oral canc.}
 & \texttt{study\_char.}    & $-0.010$ & $+0.015$ & $+0.018$ \\
 & \texttt{patient\_pop.}   & $+0.048$ & $+0.029$ & $+0.005$ \\
 & \texttt{reference\_std.} & $+0.049$ & $+0.015$ & $+0.071$ \\
 & \texttt{index\_test}     & $\mathbf{+0.476}$ & $\mathbf{+0.489}$ & $\mathbf{+0.433}$ \\
\bottomrule
\end{tabular}
\caption{Specification ablation, $\Delta F_1$\,=\,full\,$-$\,description-only,
macro-F1, per form across all three model families. Positive favors the full
specification; bold marks the oral-cancer \texttt{index\_test} collapse,
reproduced by all three models.}
\label{tab:specfull}
\end{table}

\section{Decomposition Ablation}
\label{app:decompfull}

Table~\ref{tab:decomp} gives the full corpus-by-modality breakdown (table
vs.\ scalar forms, averaged over all three model families) behind the
robustness finding in \S\ref{sec:results-scorecard}.

\begin{table}[H]
\centering
\small
\setlength{\tabcolsep}{5pt}
\begin{tabular}{lcc}
\toprule
Corpus & Table $\Delta F_1$ & Scalar $\Delta F_1$ \\
\midrule
Oral cancer   & $+0.027$ & $+0.003$ \\
Antibiotic    & $-0.036$ & $-0.012$ \\
Periodontitis & $-0.002$ & $-0.012$ \\
Ibuprofen     & $\phantom{+}0.000$ & $+0.003$ \\
\bottomrule
\end{tabular}
\caption{Decomposition ablation: overall $\Delta F_1$ (full multi-signature
pipeline minus single-prompt arm), split by table and scalar forms and
averaged over Claude, GPT, and Gemini. Positive favors the full pipeline.}
\label{tab:decomp}
\end{table}

\section{Detailed Error Analysis}
\label{app:errorfull}

Table~\ref{tab:errorfull} expands the taxonomy in
Table~\ref{tab:errorprofile}, giving for each discrepancy type what it is and
why it occurs.

\begin{figure*}[t]
  \centering
  \includegraphics[width=0.75\textwidth]{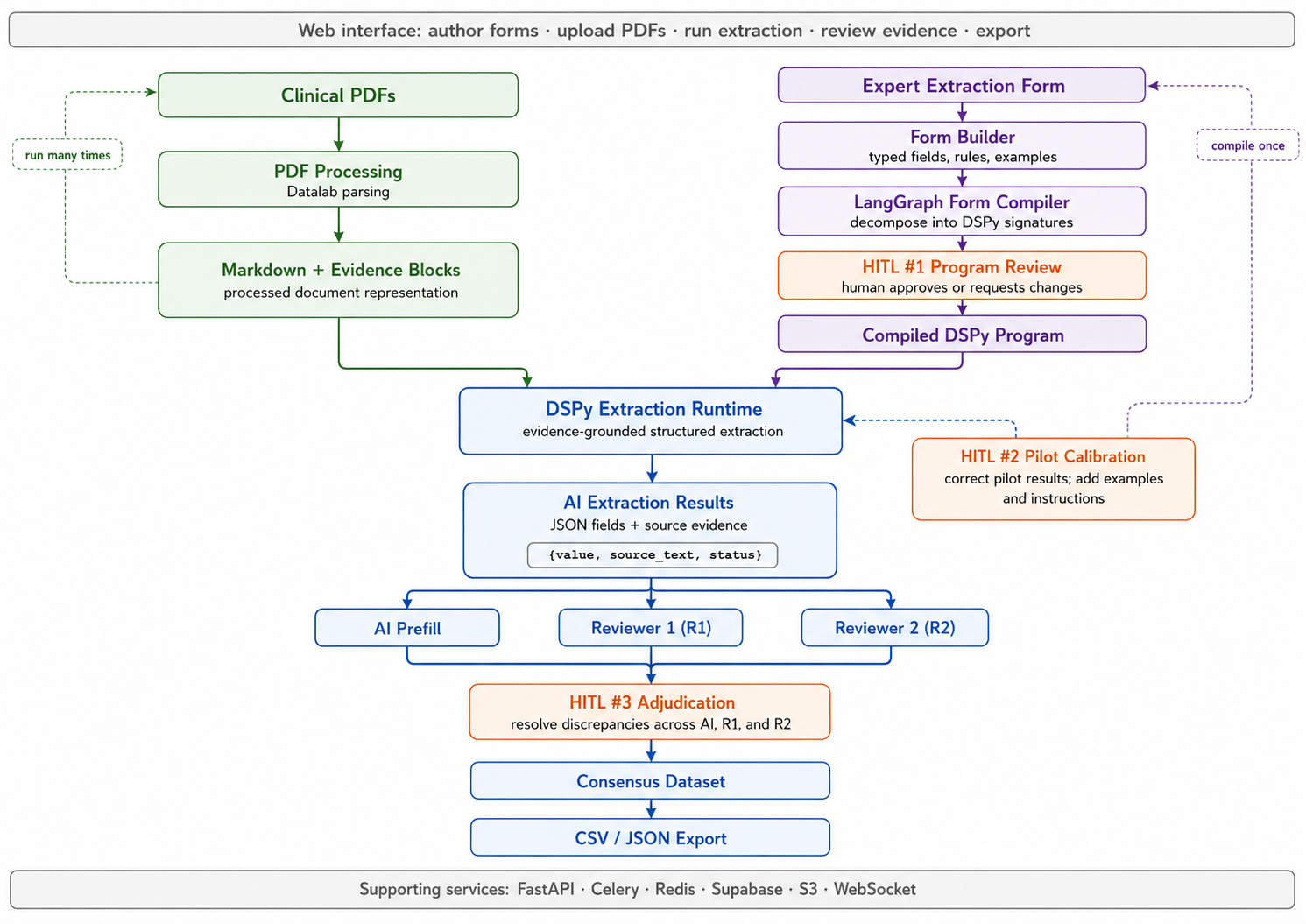}
  \captionof{figure}{The \sys pipeline. Human-in-the-loop (HITL) review enters at
  three key stages: the extraction \emph{program} (HITL~\#1), its \emph{specification}
  (HITL~\#2), and its \emph{predictions} (HITL~\#3).}
  \label{fig:architecture}
  \vspace{10pt}
  \input{sections/error_table_detail.tex}
\end{figure*}

%% file: sections/error_table_detail.tex
\small
\setlength{\tabcolsep}{4pt}
\renewcommand{\arraystretch}{1.15}
\begin{tabular}{@{}p{2.75cm} p{6.55cm} p{6.05cm}@{}}
\toprule
Discrepancy type & What it is & Why it occurs \\
\midrule
Methodological or evaluation-side\newline{\itshape 683/1003 (68\%)} & Four patterns: the ground-truth cell is wrong, blank, or falsely coded NR (27\%); a report-silent reviewer convention (back-calculated SD, converted dose, figure read off a graph) vs.\ the literal value (24\%); a field with two defensible readings and no multi-arm tie-break (11\%); or a surface-form mismatch, e.g.\ explained vs.\ bare NR (6\%). & Ground truth carries transcription slips and undocumented conventions the extraction is deliberately not allowed to infer; specifications do not disambiguate every multi-arm edge case; and scoring does not normalize every equivalent string. \\
\addlinespace[2pt]
\midrule
\multicolumn{3}{@{}l}{\textbf{Genuine model errors}\;\textbf{298/1003 (30\%)}}\\
\addlinespace[2pt]
Missing record or outcome row\newline{\itshape 75/1003 (7\%)} & A whole row that should appear in a multi-arm outcome table (a subgroup, a secondary time point, or a prose-only outcome) is absent, not merely mis-valued. & Dense multi-arm, multi-time-point tables (and prose-only outcomes) are harder for row enumeration than single-value fields. \\
\addlinespace[2pt]
Stated-value omission (false NR)\newline{\itshape 63/1003 (6\%)} & NR was coded for an event count or arm size stated plainly in the report text or a results-table cell, needing no calculation. & The value sits inside dense multi-arm results prose or a table cell the extraction did not check before defaulting to NR. \\
\addlinespace[2pt]
Wrong arm, table, or time point\newline{\itshape 60/1003 (6\%)} & A value was produced but taken from the wrong arm, wrong table, or wrong time-point bin (or a plain transcription slip). & Multiple similar-looking tables, subgroups, and time points sit close together in these results sections. \\
\addlinespace[2pt]
Cohort or adjacent-column confusion\newline{\itshape 49/1003 (5\%)} & A value taken from the whole-trial or a nested sub-cohort N instead of the arm's analysed N, or from an adjacent time-point column. & Multi-stage randomization and adjacent per-time-point columns in dense baseline tables invite cohort and column mix-ups. \\
\addlinespace[2pt]
Denominator or document-context error\newline{\itshape 37/1003 (4\%)} & A stated value was dropped, the wrong patient/lesion denominator was assigned, or generic Discussion text was reported as the study's own Methods. & Diagnostic-accuracy papers report several similar counts (received vs.\ analysed, per-arm vs.\ per-technique); citation-heavy Discussion can read like procedure. \\
\addlinespace[2pt]
Arm or regimen conflation\newline{\itshape 14/1003 (1\%)} & The wrong arm's dose, regimen, or label was copied onto a cell (e.g.\ the placebo arm described with the active arm's schedule), or a stated fact was dropped. & Multi-arm drug-versus-placebo and short- versus long-course dosing descriptions sit close together in dense prose. \\
\addlinespace[2pt]
\midrule
Uncertain/unadjudicable\newline{\itshape 22/1003 (2\%)} & A disagreement the reviewer could not confidently attribute to either a genuine model error or a methodological/ground-truth issue from the paper text alone. & The source paper does not contain enough information to adjudicate the disagreement either way, so it is left unclassified rather than forced into a category. \\
\bottomrule
\end{tabular}
\captionof{table}{Detailed error definitions and causes for the discrepancy types in
Table~\ref{tab:errorprofile} (counts and shares are given there).}
\label{tab:errorfull}